\documentclass[11pt]{article}

\usepackage[margin=1in]{geometry}
\usepackage{blindtext}
\usepackage{graphicx}
\usepackage{booktabs}
\usepackage{amsmath}
\usepackage{doi}

\title{3-D Representations for Hyperspectral Flame Tomography}

\author{
Nicolas Tricard \and
Zituo Chen \and
Sili Deng\thanks{Corresponding author: silideng@mit.edu}
}

\date{
\textit{Department of Mechanical Engineering, Massachusetts Institute of Technology \\
77 Massachusetts Ave, Cambridge, MA 02139, United States} \\
March 26, 2026
}

\usepackage{xcolor}
\usepackage{comment}

\begin{document}
\maketitle
\begin{center}
\small
\textit{This manuscript corresponds to the version presented at the 2026 Spring Technical Meeting of the Eastern States Section of the Combustion Institute (ESSCI). Extensions of this work are planned for future journal publication.}
\end{center}

\begin{abstract} 
Flame tomography is a compelling approach for extracting large amounts of data from experiments via 3-D thermochemical reconstruction.
Recent efforts employing neural-network flame representations have suggested improved reconstruction quality compared with classical tomography approaches, but a rigorous quantitative comparison with the same algorithm using a voxel-grid representation has not been conducted.
Here, we compare a classical voxel-grid representation with varying regularizers to a continuous neural representation (NN) for tomographic reconstruction of a simulated pool fire. 
The representations are constructed to give temperature and composition as a function of location, and a subsequent ray-tracing step is used to solve the radiative transfer equation to determine the spectral intensity incident on hyperspectral infrared cameras, which is then convolved with an instrument lineshape function.
We demonstrate that the voxel-grid approach with a total-variation regularizer reproduces the ground-truth synthetic flame with the highest accuracy for reduced memory intensity and runtime. Future work will explore more representations and under experimental configurations.
\end{abstract}


\section{Introduction}
Flame tomography for thermochemical state reconstruction has emerged as a leading option for experimental data procurement in the age of data-intensive combustion machine learning~\cite{DENG2025105796}.
Tomography, or 3-D flame reconstruction of fields, can be used to train chemical kinetic surrogates for predicting pollutant emissions, generate data-driven digital twins, or characterize combustion flow while circumventing error-prone and expensive computational fluid dynamics simulations by providing experimental data directly.

Flame tomography can broadly be divided into (i) an inference algorithm that maps measurements to the reconstruction and (ii) the 3-D flame representation that parametrizes and stores the thermochemical state.
Classical techniques, such as filtered backprojection (FBP) and algebraic reconstruction technique (ART), rely on linear forward models and are therefore restricted to linearized representations (\textit{e.g.}, voxel grids and linear basis expansions).
Differentiable rendering (DR), a class of methods that uses analytic gradients to iteratively reconstruct a 3-D representation via a forward operator, enables a broader range of nonlinear representations~\cite{mildenhall2020nerfrepresentingscenesneural,xie2022neuralfieldsvisualcomputing,wang20243drepresentationmethodssurvey}. This can eliminate bias from forward-model linearization, impose useful inductive constraints aiding ill-posed inversions, and enable joint end-to-end inference from parameters to measurement.
DR has provided advancements in a range of fields ranging from X-ray computed tomography~\cite{Gopalakrishnan,Grega2025} to hyperspectral imaging~\cite{chen2024hyperspectralneuralradiancefields}.
Recent work has attempted DR for flame tomography, in particular with neural-implicit representations~\cite{MOLNAR2025114298,ZHANG2023108107,Molnar_2022}.
These approaches showed that the flame may be well represented by overfitting neural networks and would not suffer from the data bias that may emerge from a more traditional pre-training/inference machine learning approach~\cite{GRAUER2023101024}.
Thus far, no existing work has rigorously compared the reconstruction quality and computational performance between various representations in DR for flame emission tomography.
Evaluating the capability of DR in combustion tomography requires determining which of the 3-D representations produces the best reconstructions at the lowest memory footprint and computational runtime.

Of the many representations used in DR pipelines, voxel grids, akin to finite volumes in computational fluid dynamics, are the most classical representation~\cite{wang20243drepresentationmethodssurvey}. 
Neural approaches have also become more prominent, where the scene is implicitly represented using a black-box model with an input of a spatial/directional query and an output of field quantities (i.e., temperature)~\cite{xie2022neuralfieldsvisualcomputing}; here, the scene is parametrized by the network's trained weights. This has the advantage of continuously representing the scene, thanks to the neural networks' continuous functional mapping from inputs to outputs. 
Additionally, neural approaches are appreciated for their ability to adaptively resolve high-resolution artifacts during training. 
The neural radiance field (NeRF), and its many variants, are a prominent example of a neural representation that has been used extensively in the field of DR~\cite{mildenhall2020nerfrepresentingscenesneural,barron2021mipnerfmultiscalerepresentationantialiasing}. 

In this research, we test regularized variants of voxel grids and neural representations in a DR pipeline for three-dimensional tomography of a simulated pool fire to reconstruct composition and temperature. For each approach, we compare its computational performance and reconstruction quality to assess which is most favorable for combustion field reconstruction.

\section{Methods}

\begin{figure}[t]
    \centering
    \includegraphics[width=1\linewidth]{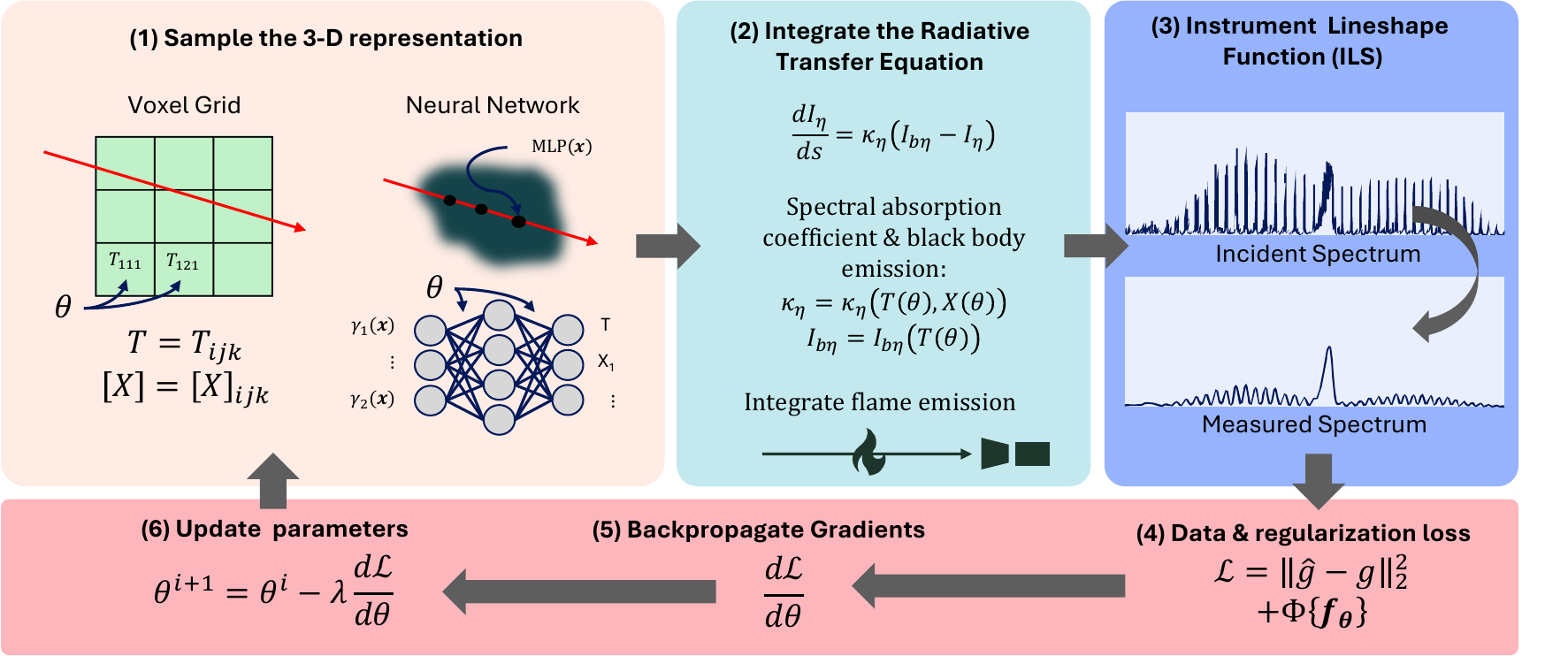}
    \caption{The tomographic process is conducted iteratively. (1) Temperature and composition are obtained by sampling the 3-D representation. We test continuous (Neural Network) and discrete (voxel) representations. (2) The radiative transfer equation is solved along each ray for spectral intensity $I_\eta(s)$, where $s$ is the path length, $\eta$ is the wavenumber, $\kappa_\eta(T,X)$ is the temperature- and composition-dependent spectral absorption coefficient, and $I_{b,\eta}(T)$ is the blackbody emission. (3) The Michelson spectral ILS (triangular apodization) is applied to the incident spectra. (4) The loss function is evaluated, (5) gradients with respect to representation parameters are computed, and (6) parameters are updated via gradient descent. This procedure is repeated until the predicted measurement $\hat{g}$ matches the ground truth $g$.
    }
                \label{fig:operation}
\end{figure}

\subsection{Preliminaries: Forward and Inverse Steps}
The DR pipeline forward model is shown in Fig.~\ref{fig:operation}. The forward operator $\mathcal{H}$ maps our 3-D field representation $f_\theta$ parameterized by $\theta$ to an observable $\hat{g}$ as $\hat{g}=\mathcal{H}[f_\theta]$.
Here, we model $\mathcal{H}$ as an instantaneous emission-Fourier transform infrared (FTIR) imaging setup, a broadband measurement device that relies on a Michelson interferometer for spectral multiplexing~\cite{FTIR}.
We seek to reconstruct temperature and composition. Thus, our 3-D representation $f_\theta$ maps a physical location $\textbf{x}$ to these values. Then, a line-by-line ray-tracing model is solved using $f_\theta$ to determine the radiative emission spectra incident on a camera~\cite{Rothman2010HITEMPDatabase}. 
Finally, we spectrally convolve the resulting radiation with an sinc-squared instrument lineshape (ILS) function to obtain the measurement $\hat{g}$~\cite{FTIR}. We repeat this process for every ray cast through the scene, where rays propagate from the camera origins through pixel centerpoints into the scene.

The inverse model is to obtain the underlying parameters $\theta$ from the measurement $g$, as $f_\theta=\mathcal{H}^{-1}[g]$. In doing so, we obtain a 3-D reconstruction of thermochemical states. The parameters are determined by minimizing a loss function $\mathcal{L}$, defining the deviation between $g$ and our predicted measurement $\hat{g}$. Here, we apply L-2 loss,
\begin{equation}
    \mathcal{L}=||\mathcal{H}[f_\theta]-g||^2_2+\lambda_l\Phi\{f_\theta\}
    \label{eq:loss}
\end{equation}
for regularization parameter $\lambda_l$ (assigned to 0.001), and regularizer $\Phi$, which is defined for no regularization as $\Phi=0$, Tikhonov as $\Phi=||\mathbf{\nabla f_\theta}||^2_2$, and total variation as $\Phi=||\mathbf{\nabla{}f_\theta}||_1$, for normalized field quantities derived from $f_\theta$. 
For voxel grids, these regularizers are evaluated on the normalized temperature and species fields using first-order forward finite differences in each Cartesian direction,
\begin{equation}
L=
\begin{bmatrix}
\Delta x^{-1}D_x \\
\Delta y^{-1}D_y \\
\Delta z^{-1}D_z
\end{bmatrix},
\qquad
(D_x f)_{i,j,k}=f_{i+1,j,k}-f_{i,j,k},
\end{equation}
with analogous definitions for $D_y$ and $D_z$. Thus, the total variation penalty is the sum of the mean absolute directional derivatives, while the Tikhonov penalty is the sum of the mean squared directional derivatives. For neural representation, we use autodifferentiation on the implicit field evaluated at sampled voxel-center coordinates to compute these spatial derivatives directly, avoiding the construction of a full voxel grid during optimization.
The loss is minimized using the iterative gradient descent $\theta^{i+1}=\theta^i-\lambda{}\frac{d\mathcal{L}}{d\theta}$, for iteration $i$ and learning rate $\lambda$, where $\frac{d\mathcal{L}}{d\theta}$ is obtained through autodifferentiation of the forward pipeline. 

\subsection{Representations}

\vspace{5pt}\noindent
\textit{Voxel Grid:}
Voxel grids (VGs) are a standard control-volume-based approach to 3-D field representation in which the domain is decomposed into a finite number of contiguous axes-aligned boxes, each with uniform field properties.
The ray tracing procedure is performed in the domain via light backtracking and classical ray tracing integration~\cite{Modest2022ChapterMediac} through the piecewise homogeneous medium using 3-D Amanatides and Woo algorithm~\cite{3D_DDA}.
The unknown parameters are the temperature and species mole fractions in each voxel.
We initialize the voxel grid using the ground truth tested voxel field (of identical resolution) and perturb each voxel with Gaussian random noise of standard deviation 20\% of the ground truth value.

\vspace{5pt}\noindent
\textit{Neural Implicit Field:}
Unlike voxel grids, where parameters directly correspond to locations in 3-D space, a neural representation uses a neural network to represent the scene implicitly. 
Our approach is inspired by Neural Radiance Fields (NeRF)~\cite{mildenhall2020nerfrepresentingscenesneural}, but with two major differences. (1) Classical NeRF maps position and direction to color intensity and extinction coefficient $(\mathbf{x}, \mathbf{d}) \mapsto (c, \sigma)$, but we assume that emission from the flame is isotropic and so we drop the ray-direction input to the neural network. (2) We apply infrared hyperspectral volume rendering instead of RGB, adding a spectral forward model mapping T, X, to spectral opacity, and producing a high-resolution emission spectrum incident to the camera. 

This results in the neural implicit field:
\begin{equation}
    [T,X]=f_\theta(\gamma(\textbf{x}))
\end{equation}
for multi-layer perceptron $f_\theta$ and sinusoidal positional encoding $\gamma$ as defined in Ref.~\cite{mildenhall2020nerfrepresentingscenesneural}.
We use 64 hidden dimensions, 4 hidden layers, and 10 positional encoding frequencies.
Initialization is performed overfitting the neural network to the same 3-D field used in the voxel grid initial guess procedure.

In implementation, we use two MLP branches, a coarse network and a fine network, each queried only by encoded 3-D position. 
During rendering, rays are first intersected with the scene's bounding box to determine near and far bounds. An initial pass then performs stratified sampling along each ray, evaluates $f_\theta$ using the coarse MLP at the sampled points to predict $T$ and $X$, and constructs sampling weights that concentrate effort in regions of high expected emission and absorption. Unlike in the original NeRF~\cite{mildenhall2020nerfrepresentingscenesneural} where importance sampling is based on density-derived opacity, $\alpha=1-\exp(-\kappa{}\delta{})$, for single-scalar density $\kappa$ and opacity $\alpha$, the absorption coefficient in our case varies with wavenumber. To obtain a scalar quantity for sampling, we select the Planck-mean absorption coefficient to conduct this importance sampling routine, defined as 
\begin{equation}
\kappa = \frac{\int_0^\infty \kappa_\eta I_{b\eta} d\eta}{\int_0^\infty I_{b\eta} d\eta}~,
\end{equation}
which enables thermodynamically consistent weighting across the spectrum.
A second fine pass resamples along the ray using this importance distribution, re-evaluates the implicit field at the merged sample set, and performs volume rendering on the resulting $(T,X)$ profiles. Both coarse and fine network weights are learned during this procedure. This hierarchical ray-tracing procedure preserves the continuous spatial representation of NeRF while adapting it to infrared hyperspectral flame imaging.

\begin{figure}[t]
    \centering
    \includegraphics[width=\linewidth]{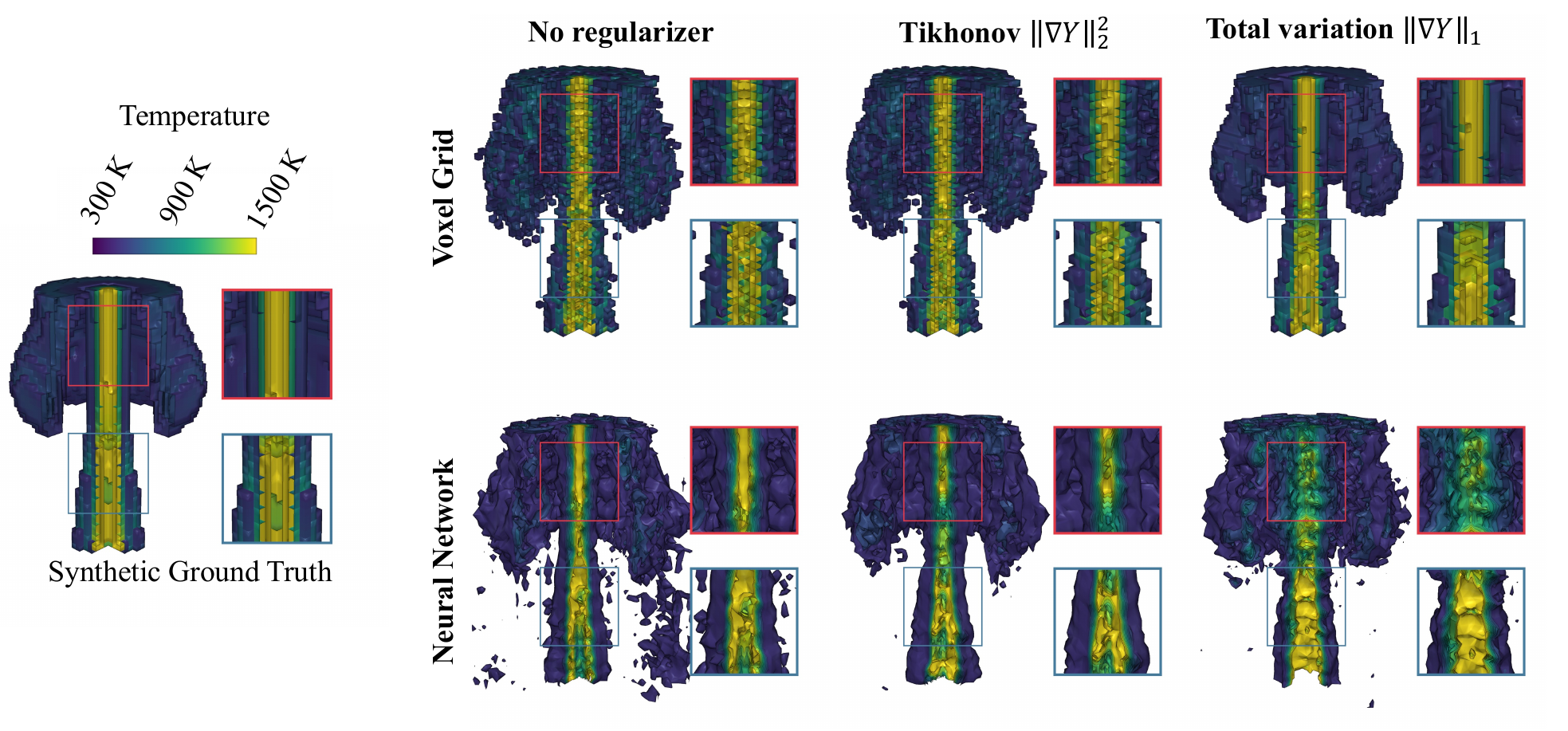}
    \caption{Cutaway plots of 3-D reconstructions from the DR pipeline against the ground truth. Isocontours of temperature are shown for intervals of 100 K between 500 and 1500 K.}
    \label{fig:Fig2}
\end{figure}

\section{Results and Discussion}

We test DR on a turbulent pool fire solved using \texttt{OpenFOAM 5.x}~\cite{Weller1998TensorialContinuum}. This fire consists of a 20 cm diameter CH$_4$ pool with a constant-flow injection rate of 0.01 meters per second into a 1x1x1 meter cell domain discretized into 216,000 cells. 
We extract a single timestep at $t=$0.9 s and attempt reconstruction of temperature and CH$_4$, CO$_2$, H$_2$O,  using four cameras at $\pm1.5$ meter x and $\pm1.5$ meter y centered origins. Each camera produces a 32x32 pixel image with a focal length of 0.59 m. Each ray then consists of a 650 cm$^{-1}$ to 725 cm$^{-1}$ spectral range at a line-by-line resolution of 0.04 cm$^{-1}$ and FTIR-convolved resolution of 8 cm$^{-1}$.

A visual comparison of temperature isocontours is presented in Fig.~\ref{fig:Fig2} alongside the ground truth. The representations each reconstruct the general structure of the flame, albeit with their own respective visualization caveats: voxel grids present discrete voxel chunks, and NeRF requires probing discrete points (we choose to probe at the ground truth voxel-grid centerpoints) to extract 3-D varying thermochemistry.
The neural network approach, thanks to its continuous representation, can provide non-linear interpolation between voxel grid points and thus smoothly captures details such as sharp temperature gradients around the flame centerline.
However, it suffers from more spurious artifacts than VG despite regularization, arising due to the difficulty of fitting a NN of limited size.

We also report in Table~\ref{tab:psnr_comparison} the 3-D mean square error,
$\mathrm{MSE}=\left\lVert f_\theta(\mathbf{x}_{\mathrm{VG}}) - f_{\mathrm{gt}}(\mathbf{x}_{\mathrm{VG}}) \right\rVert_2^2/N$, evaluated for $N$ voxels at their centerpoints $\mathbf{x}_{\mathrm{VG}}$,  and fields are normalized to their minimum and maximum values.
Between the representations VG and NN, and the regularization options of no regularizer (NR), Tikhonov (Tikh.), and total variation (TV), we see that the VG with TV combination performs best at reconstructing the flame after 2000 epochs.
While TV works best with VG, Tikh. is the best performing regularization approach with NN. The difference arises because NNs already contain numerous inductive biases arising from their propensity to fit low-frequency signals. VG, meanwhile, contains no built-in biases and requires a stronger prior.
For both representations, employing no regularization results in excessive noise due to the problem's ill-posed nature.

The comparison of field reconstruction qualities in Table~\ref{tab:psnr_comparison} shows superior reconstruction quality of temperature over mole fraction, with H$_2$O reconstructed at the lowest quality, likely due to its minimal contribution to the overall flame emission compared to CO$_2$.

\begin{table}[t]
\centering
\caption{Performance comparison of different 3-D representations for the 3-D pool fire.}
\label{tab:psnr_comparison}
\begin{tabular}{lccccc}
\toprule
\textbf{Method} & Epoch Time [ms] & Mem. [MB] & \textbf{MSE$_T$} & \textbf{MSE$_{CO_2}$} & \textbf{MSE$_{H_2O}$} \\
\midrule

VG/NR & 430 & 3.0 & 0.0295 & 0.0436 & 0.0416 \\
VG/Tikh. & 432 & 3.0 & 0.0203 & 0.0391 & 0.0384 \\
VG/TV & 431 & 3.0 & \textbf{0.0121} & \textbf{0.0219} & \textbf{0.0197} \\

\midrule

NN/NR & 479 & 65.0 & 0.0571 & 0.0871 & 0.2569 \\
NN/Tikh. & 489 & 66.0 & \textbf{0.0268} & \textbf{0.0705} & \textbf{0.0946} \\
NN/TV & 493 & 66.0 & 0.0754 & 0.1427 & 0.4356 \\

\bottomrule
\end{tabular}
\end{table}

\section{Conclusions}

We apply differentiable rendering with voxel grids and neural implicit fields and compare their performance for hyperspectral infrared flame emission tomography. We formulate the non-linear inverse problem as an iteration over a differentiable forward operator, mapping the scene representation parameters to 2-D FTIR images situated around the flame. The voxel grid representation with a total-variation regularization penalty performed best both qualitatively and quantitatively as measured by mean square error.

We note this approach is limited in numerous ways. (1) Our ground-truth flow-field is synthetic and may not contain all of the flame eddies or bulk structures seen in real flames. (2) The imaging forward model is based on a concept of a highly-expensive and well-calibrated hyperspectral imaging system with no noise. (3) We neglect soot emission from the pool flame. Future work will implement this tomographic approach under real conditions seen in experiment.

\section{Acknowledgements}
We acknowledge the funding support from ExxonMobil Corporation and the Carbon
Hub and the Kavli Foundation Exploration Award in Nanoscience for
Sustainability LS-2023-GR-51-2857. In addition, NT thanks the  G.E. Vernova for PhD funding.
We also thank the National Laboratory of the Rockies for GPU computing resources.

\bibliography{references}

\end{document}